\documentclass[runningheads]{llncs}
\usepackage{graphicx}
\usepackage[colorlinks=true,allcolors=blue]{hyperref}
\usepackage{amsmath,amssymb,amsfonts}

\usepackage{multirow}
\usepackage{booktabs}
\usepackage{placeins}
\usepackage{bm}
\usepackage{subfigure}
\usepackage[misc]{ifsym} 
\begin{document}
\title{Transformer based multiple instance learning for weakly supervised histopathology image segmentation
}
\titlerunning{Swin Transformer Based MIL}
%
\author{Ziniu Qian\inst{1}, Kailu Li\inst{1}, Maode Lai\inst{2,3}, Eric I-Chao Chang\inst{4}, Bingzheng Wei\inst{5}, Yubo Fan\inst{1}, Yan Xu\inst{1}$^{(\textrm{\Letter})}$}
\authorrunning{Qian et al.}

\institute{
School of Biological Science and Medical Engineering, State Key Laboratory of Software Development Environment, Key Laboratory of Biomechanics, Mechanobiology of Ministry of Education and Beijing Advanced Innovation Centre for Biomedical Engineering, Beihang University, Beijing 100191, China\\
\email{xuyan04@gmail.com} \and
China Pharmaceutical University, Nanjing 210009, China \and
Zhejiang University, Hangzhou 310058, China \and
Microsoft Research, Beijing 100080, China \and
Xiaomi Corporation, Beijing 100085, China
}
\renewcommand{\thefootnote}{}
\footnotetext{$^*$Ziniu Qian and Kailu Li are contributed equally to this work.}
\maketitle              

\begin{abstract}
Hispathological image segmentation algorithms play a critical role in computer aided diagnosis technology. The development of weakly supervised segmentation algorithm alleviates the problem of medical image annotation that it is time-consuming and labor-intensive. As a subset of weakly supervised learning, Multiple Instance Learning (MIL) has been proven to be effective in segmentation. However, there is a lack of related information between instances in MIL, which limits the further improvement of segmentation performance. In this paper, we propose a novel weakly supervised method for pixel-level segmentation in histopathology images, which introduces Transformer into the MIL framework to capture global or long-range dependencies. The multi-head self-attention in the Transformer establishes the relationship between instances, which solves the shortcoming that instances are independent of each other in MIL. In addition, deep supervision is introduced to overcome the limitation of annotations in weakly supervised methods and make the better utilization of hierarchical information. The state-of-the-art results on the colon cancer dataset demonstrate the superiority of the proposed method compared with other weakly supervised methods. It is worth believing that there is a potential of our approach for various applications in medical images.
\end{abstract}
\section{Introduction}
Histopathology image segmentation is of great significance in computer-assisted diagnostics (CAD), which assists doctors quickly trace abnormal tissue areas. Recently, effective supervised methods always rely on a large amount of high-quality annotations \cite{ronneberger2015u,chen2016dcan,xing2016transfer}. However, it is time-consuming and labor-intensive for doctors to manually label regions of interest (ROIs) in histopathology images \cite{yu2020weakly}. Therefore, there is a strong requirement of automated methods to segment and trace ROIs in histopathology images with weakly supervised methods. 
In this paper, we consider the design of weakly supervised segmentation methods with image-level annotations for their low cost and wide applicability \cite{zhou2018brief}. 

Multiple instance learning (MIL) \cite{dietterich1997solving} is a subset of weakly supervised methods, which has demonstrated its effectiveness on segmentation tasks in previous studies \cite{xu2012multiple,xu2014weakly,jia2017constrained}. Training datasets of MIL are set as several bags that contain multiple instances. The available labels are only assigned at the bag-level. MIL is able to predict instance-level labels except performing bag-level classification. In this work, cancerous images or not are regarded as bags, while pixels in images are regarded as instances. Labels of pixels can be predicted by MIL with the annotations of images so that semantic segmentation can be performed. However, MIL methods are all limited by the fact that the instances within the bags are independent of each other. The fact leads to a lack of relationships between instances with similar contextual information. 

Transformer \cite{vaswani2017attention} has demonstrated its excellent performance both on Natural Language Process (NLP) and Computer Vision (CV). Different from convolutional neural networks (CNNs) that focus on the local receptive field at each convolution layer, Transformer can capture global or long-range dependencies with a self-attention mechanism \cite{chen2021transunet}.  Self-attention mechanism computes the response at a position in a sequence by attending to all positions and taking their weighted average in an embedding space. That is, self-attention mechanism aggregates contextual information from other instances in a bag in MIL. By weighting value with attention matrix, self-attention mechanism increases the difference between classes, which is the distance between foreground and background in semantic segmentation. Therefore, the feature maps from Transformer implicitly include relationships between instances in MIL. In previous studies, several researchers have proposed MIL methods integrated with Transformer that achieve superior performance over CNNs models \cite{li2021dt,yu2021mil,shao2021transmil}. However, existing studies in this field mostly focus on image classification, rarely addressing the challenge of segmentation. To the best of our knowledge, we are the first to attack semantic segmentation on histopathology images using Transformer combined with MIL. 

Among the developed Transformer methods, Swin Transformer \cite{liu2021swin} constructs multi-scale feature maps and achieves state-of-the-art performance on several vision tasks. It is obvious that algorithms utilizing hierarchical information from multi-scale feature maps show better performance \cite{yi2019multi}. On the other hand, Swin Transformer provides higher resolution feature maps than other Transformer methods, which is benefit to facilitate prediction maps for lower upsampling ratio. Therefore, we explore adapting Swin Transformer in the MIL framework. 
In addition, it is difficult to constrain the learning process in the MIL method due to a lack of supervision. On the other hand, Transformer relies more on large datasets than CNNs methods, which are difficult to obtain in medical image analysis. To address these problems, we introduce deep supervision \cite{jia2017constrained,lee2015deeply} to make better use of image-level annotations and strengthen constraints.

In this work, we introduce the Transformer into the MIL framework to perform histopathology image segmentation for the first time. We propose a novel MIL method leveraging Swin Transformer encoder, decoder and deep supervision to build a trainable bag embedding module for generating prediction masks. Swin Transformer encoder learns deep feature representations, where attention weights are assigned to features. To extend constraints, we introduce deep supervision by producing multi-scale side-outputs from the encoder which can be leveraged adequately by a fusion layer. Our code will be released on https://github.com/Nexuslkl/Swin\_MIL. 
The contributions can be summarized as follows:
\begin{itemize}
\item We propose a novel Transformer based method for weakly supervised semantic segmentation. Deep supervision is leveraged in the framework to utilize multi-scale features and provide more supervised information.
\item We explore the combination of Transformer and MIL for weakly supervised segmentation on histopathology images, which, to our best knowledge, is the first attempt at this task.
\item Multi-head self-attention of Transformer builds long-range dependencies to solve the problem that instances in MIL are independent of each other.
\end{itemize}

\section{Method} 
In this paper, the motivation is to introduce Transformer to solve the shortcoming of independent instances in MIL for histopathology image segmentation.
The proposed Swin-MIL consists of Swin Transformer encoder, decoder and deep supervision. Swin Transformer encoder plays a role in assigning attention weights to features. By capturing global information, the self-attention module effectively highlights features that have better interpretability at each stage. Decoder of each stage generates pixel-level predictions as side-outputs, where the final segmentation maps are fused with side-outputs across all stages. Deep supervision is to constrain the training process with only image-level annotations. An overall framework of the proposed Swin-MIL is illustrated in \autoref{fig1}.

\subsubsection{Definition:} Let $S = \left\{(X_n, Y_n),n = 1, 2, 3,..., m \right\}$ donates our training set, where $X_n \in R^{H \times W}$ denotes the $n$th input image and $Y_n \in \left\{0, 1\right\}$ refers to the image-level label assigned to the $n$th input image. Here $Y_n = 1$ refers to a positive image and $Y_n = 0$ refers to a negative image.

\subsubsection{Swin Transformer encoder:} MIL does not perform as well as fully supervised semantic segmentation methods. There is a problem that instances in the standard MIL maintain their independencies, which is in conflict with a characteristic named category consistency in semantic segmentation tasks. It means that pixels in the same class should have similar features, while those in a separate class should have dissimilar features \cite{huang2019ccnet}. To address the shortcoming of MIL, Swin Transformer encoder is introduced that incorporates the self-attention mechanism into the MIL setting. By introducing a shifted window partitioning approach which alternates between two partitioning configurations in consecutive Swin Transformer blocks, Swin Transformer encoder builds relationships between long-range tokens with an efficient computation. Therefore, similar features get high attention weights while dissimilar ones get low attention weights, which leads to an improvement in distinguishing foreground and background.

The image $X_n$ is cropped into non-overlapping patches with patch size of $4 \times 4$ firstly. Each patch is treated as a “Token” and fed into several Swin Transformer blocks, which enhance the significant regions in feature maps and inhibit the influence of irrelevant regions through multi-head self-attention. The Swin Transformer blocks refer to as “Stage 1” together with a linear embedding layer, which projects the raw-valued feature to an arbitrary dimension. To produce a multi-scale feature representation, patch merging layers are introduced to reduce the number of tokens as the network gets deeper. The patch merging layers and several Swin Transformer blocks are denoted as “Stage 2” and “Stage 3”, which output feature maps with resolution of $\frac{H}{8} \times \frac{W}{8}$ and $\frac{H}{16} \times \frac{W}{16}$, respectively. It is evident in our experiments that the backbone with three stages is sufficiently powerful to extract features. Then the output feature maps are fed into decoder.

\begin{figure*}[!t]
\centering
\includegraphics[width=\textwidth]{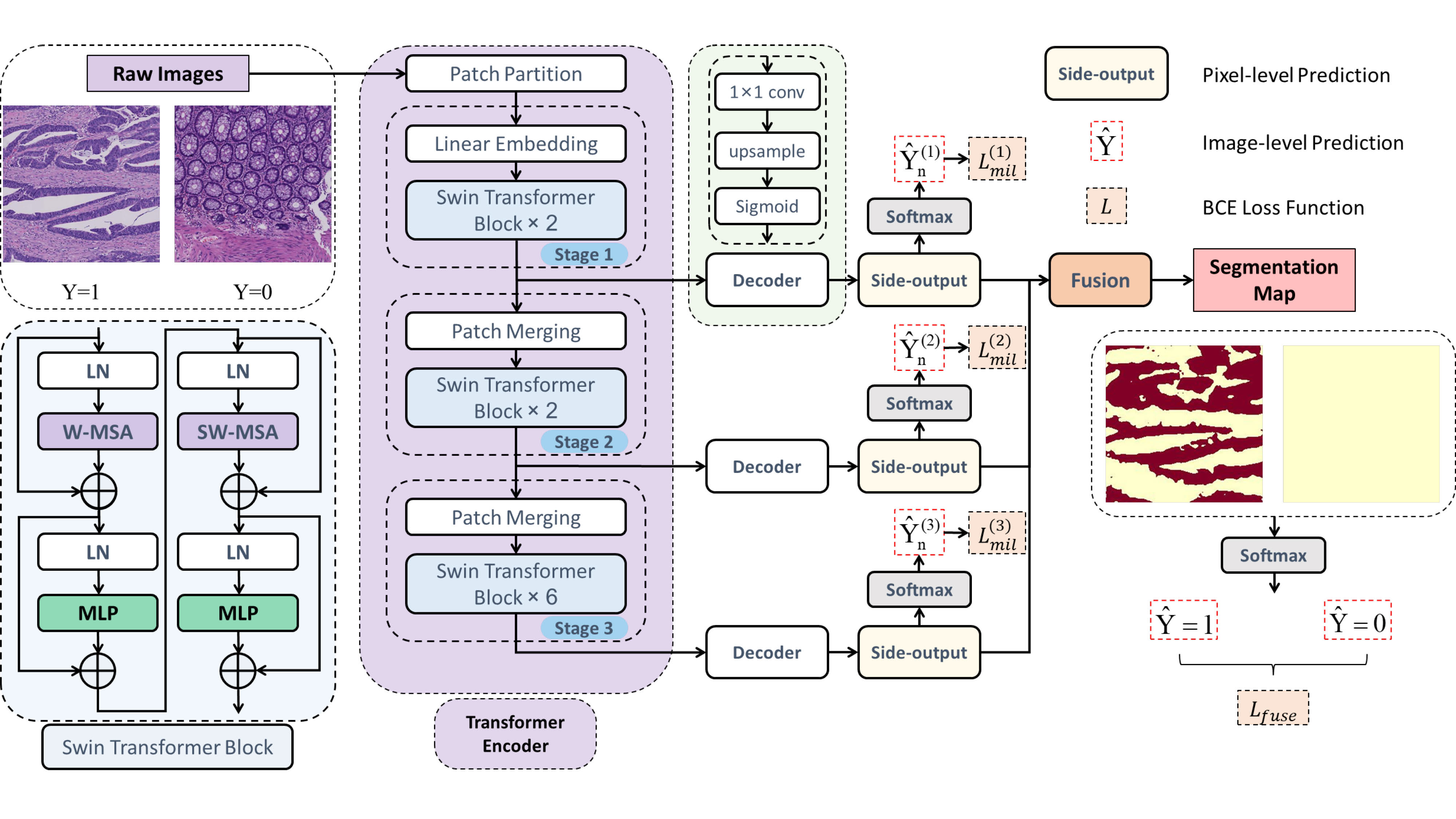}
\caption{An overview of our method. Under the MIL setting, we adopt the first three stages of the Swin Transformer encoder. At each stage, the deep supervision layer produces side-outputs which can be seen as predictions. In addition, a fusion layer is proposed to adequately leverage the multi-scale predictions across all side-outputs.}
\label{fig1}
\end{figure*}

\subsubsection{Decoder:} The channel size of the output feature map is squeezed to 1 by a $1 \times 1$ convolutional layer. After a bilinear upsampling operation, it is restored to the original size ($H \times W$) and finally activated by the sigmoid to generate a predicted probabilistic map as side-outputs. A fusion layer is proposed to adequately leverage the multi-scale side-outputs across all stages to predict final segmentation maps. 
We can predict the classification of the image by each instance in the bag. $\hat{Y}_n(i,j)$ is denoted as the probability of the pixel $p_{ij}$ in the $n$th image, where $(i,j)$ denotes the position of the pixel $p_{ij}$ in $X_n$. Therefore, a softmax function is often used to replace the hard maximum approach. Generalized Mean (GM) \cite{zhang2005multiple} is utilized as our softmax function, which is defined as:
\begin{equation}
    \hat{Y}_n=\left( \frac{1}{\left| X_n \right|}\sum_{i,j}\left [\hat{Y}_n(i,j) \right]^r \right)^\frac{1}{r},
\end{equation}
where the parameter $r$ controls the sharpness and proximity to the hard function: $ \hat{Y}_n \rightarrow \mathop{max}\limits_{i,j} \hat{Y}_n(i,j)$ as $r \rightarrow \infty $. 

\subsubsection{Deep supervision:} To address the lack of supervision information in MIL and utilize hierarchical information effectively, we introduce deep supervision. Our goal is to train the model by minimizing the loss between output predictions and ground truths, which is designed in the form of the cross-entropy loss function as follows:
\begin{equation}
    L_{mil}^{(t)}=-\sum_n\left( \textbf{I}\left( Y_n=1 \right)log\hat{Y}_n^{(t)}+\textbf{I}\left( Y_n=0 \right)log(1-\hat{Y}_n^{(t)}) \right),
\end{equation}
\begin{equation}  
    L_{fuse}=-\sum_n( \textbf{I}\left( Y_n=1 \right)log\hat{Y}_{n,fuse}+\textbf{I}\left( Y_n=0 \right)log(1-\hat{Y}_{n,fuse})),
\end{equation}
where $t$ is denoted to the number of stage, $\hat{Y}_n^{(t)}$ and $\hat{Y}_{n,fuse}$ are calculate by Equation (1), and $\textbf{I} (\cdot)$ is an indicator function.

\subsubsection{Loss function:}
The final objective loss function is defined as below:
\begin{equation}
    L=\sum_{t=1}^3 L_{mil}^{(t)}+L_{fuse}.
\end{equation}

In each side-output and fusion layer, the loss function is computed in the form of deep supervision without any additional pixel-level label supervision. After that, the parameters are learned by minimizing the objective function via backpropagation using the stochastic gradient descent algorithm.

\section{Experiment}
\subsection{Dataset}
A hispathological tissue dataset was selected to evaluate the effectiveness of our approach. The dataset is a Haematoxylin and Eosin (H\&E) stained histopathology image dataset of colon cancer reported by Jia et al. \cite{jia2017constrained}, which consists of 330 cancer (positive) and 580 non-cancer (negative) images. In this dataset, 250 positive and 500 negative images were used for training; 80 cancer and 80 non-cancer images were used for testing. These images were obtained from the NanoZoomer 2.0HT digital slide scanner produced by Hamamatsu Photonics with a magnification factor of 40, i.e. 226 nm/pixel. Each image has a resolution of 3, 000 × 3, 000. Due to memory limits, the original 3, 000 × 3, 000 pixels can not be loaded directly. Thus, in all experiments, images were resized to 256 × 256 pixels. For simplicity, we used Pos to refer to cancer images and Neg to refer to non-cancer images. The ground truths are pixel-level annotations which were provided by two pathologists. (1) If the overlap between the two cancerous regions labeled by the two pathologists is larger than 80$\%$, we use the intersection of the two regions as the ground truth; (2) if the overlap is less than 80$\%$, or if a cancerous region is annotated by one pathologist but ignored by another, a third senior pathologist will step in to help determine whether the region is cancerous or not.
\subsection{Implementation} 
All experiments were implemented on the PyTorch framework and conducted on NVIDIA GeForce RTX 3090 GPUs with 24G memory. For training, the Adam optimizer is employed to train the model with a weight decay of 5e-4 and a fixed learning rate of 1e-6. The batch size is set to 4 per GPU, with an epoch of 60. The parameter r of the generalized mean function is set to 4. Learnable weights will cause over-segmentation problems and degrade performance. With the limited supervision of image-level annotation, MIL methods with learnable weights has a tendency to perform better classification but not segmentation. Thus, we adopt a strategy, using fixed weights, to preserve multi-scale information. The fixed weight values we set are optimal settings determined by experiments. Weights of 0.3, 0.4, 0.3 are selected for the three side-output layers. The learning rates of the side-output layers are set to 1/100 of the global learning rate. We used the pretrained model of Swin-T from ImageNet on our backbone and Xavier initialization \cite{glorot2010understanding} to initialize the side-output layers.

F1-score (F1) and Hausdorff Distance (HD) are employed to evaluate the quality of semantic segmentation and the boundary of prediction mask, respectively. Here, the F1-score in positive images is equivalent to the dice coefficient.

\begin{figure*}[!t]
\centering
\includegraphics[width=\textwidth]{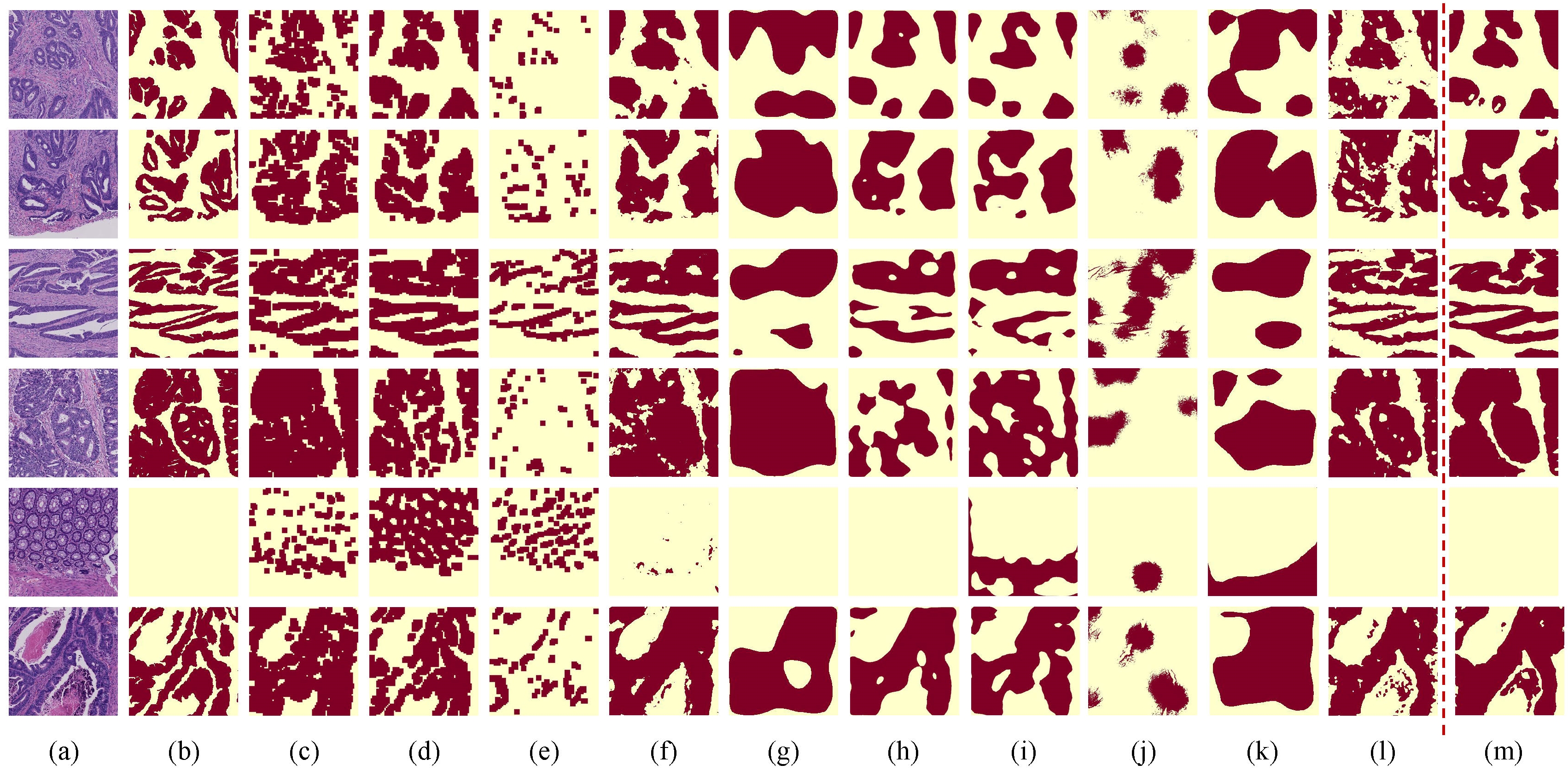}
\caption{Visualization results of all methods. (a) Raw
Image. (b) Ground Truth. (c) DA-MIL. (d) DeepAttnMISL. (e) GA-MIL. (f) DWS-MIL. (g) OAA. (h) MDC-UNet. (i) MDC-CAM. (j) PRM. (k) VGG-CAM. (l) Ours. (m) U-Net. The rows 1-4 are the results on positive images. The row 5 are the results on negative images. The row 6 are the failure cases.}
\label{fig2}
\end{figure*}

\subsection{Comparisons}

\begin{table*}[!t]
\caption{Comparisons with weakly supervised methods and fully supervised methods. Pos and Neg means the results are conducted on positive images and negative images, respectively. WSL and FSL are denoted to weakly supervised learning and fully supervised learning, respectively.}
\label{tab1}
\centering
\begin{tabular}{llcccc}
\toprule
Type & Method & $\rm F1^{Pos}$ & $\rm HD^{Pos}$ & $\rm F1^{Neg}$ & \begin{tabular}[c]{@{}c@{}}Running \\ Time(s)\end{tabular} \\ \hline
\multirow{5}{*}{MIL based WSL:} & \textbf{Swin-MIL (Ours)} & \textbf{0.850} & \textbf{10.463} & \textbf{0.999} & 0.0226  \\
& DA-MIL (2020) \cite{hashimoto2020multi}& 0.791 & 12.962 & 0.755 & 0.1635 \\
& DeepAttnMISL (2020) \cite{yao2020whole} & 0.772 & 11.111 & 0.634 & 0.1389 \\ 
& GA-MIL (2018) \cite{ilse2018attention} & 0.355 & 16.289 & 0.939 & 0.1695 \\
& DWS-MIL (2017) \cite{jia2017constrained} & 0.833 & 15.179 & 0.986 & \textbf{0.0142} \\ \hline
\multirow{5}{*}{CAM based WSL:} & OAA (2021) \cite{jiang2021online}& 0.744 & 28.972 & \textbf{0.999} & 0.0190 \\ 
& MDC-UNet (2018) \cite{wei2018revisiting} & 0.744 & 17.071 & 0.998 & 0.0157 \\
& MDC-CAM (2018) \cite{wei2018revisiting}& 0.726 & 13.754 & 0.760 & 0.0167 \\
& PRM (2018) \cite{zhou2018weakly}& 0.561 & 24.468 & 0.995 & 0.9277 \\
 
& VGG-CAM (2016) \cite{zhou2016learning} & 0.675 & 23.665 & 0.645 & 0.0160 \\\hline
FSL: & U-Net (2015) \cite{ronneberger2015u} & 0.885 & 7.428 & 0.997 & 0.0153 \\ \bottomrule
\end{tabular}
\end{table*}

Attention mechanism is the core of Transformer to capture global or long-range dependencies. Hence, some attention-based MIL methods \cite{ilse2018attention,hashimoto2020multi,yao2020whole} which can achieve patch-level segmentation by visualizing the attention map with image-level annotations, were conducted for comparison. Weakly supervised segmentation methods for natural images are also reimplemented on our dataset. Besides, we reimplemented DWS-MIL\cite{jia2017constrained} framework using PyTorch instead of Caffe toolbox and introduced a batch normalization layer between convolutional layer and RELU for fair comparison. As shown in \autoref{tab1} and \autoref{fig2}, compared with recent methods \cite{jiang2021online,yao2020whole} relied on image-level annotations, our method significantly outperforms those methods (state-of-the-art) and is close to the fully-supervised U-Net \cite{ronneberger2015u}. Attention-based MIL methods mainly focus on classification tasks, where the limitation of patch-level segmentation leads to a low-quality boundary. Accurate class activation maps (CAMs) are generated CAM-based weakly supervised methods \cite{zhou2016learning,wei2018revisiting,zhou2018weakly,jiang2021online}, which are sensitive to the location of cancerous regions while ineffective in accurately segmentation. The MDC methods use dilated convolutional layers to obtain larger receptive fields, which achieve better performance than VGG-CAM. PRM and OAA can perform weakly supervised segmentation based on class activation maps to generate a complete mask prediction. However, it is obvious that these methods do not perform well in histopathology images. Because the natural image instance segmentation objects (people, vehicles, and animals) often have regular shapes and characteristics, the object regions are relatively complete. Nevertheless, the objects in the histopathology images are often of irregular shapes. It is difficult to generate accurate predictions by expanding which is sensitive to boundaries. Besides, we show the failure case in the row 6 of \autoref{fig1}. Given the complexity and the diversity of the tissue appearance, our method misses the accurate segmentation boundary under the limited supervision with image-level annotation.

\subsection{Ablation Study}
\subsubsection{Effect of different backbone:}
\begin{figure*}[!t]
\centering
\includegraphics[width=0.8\textwidth]{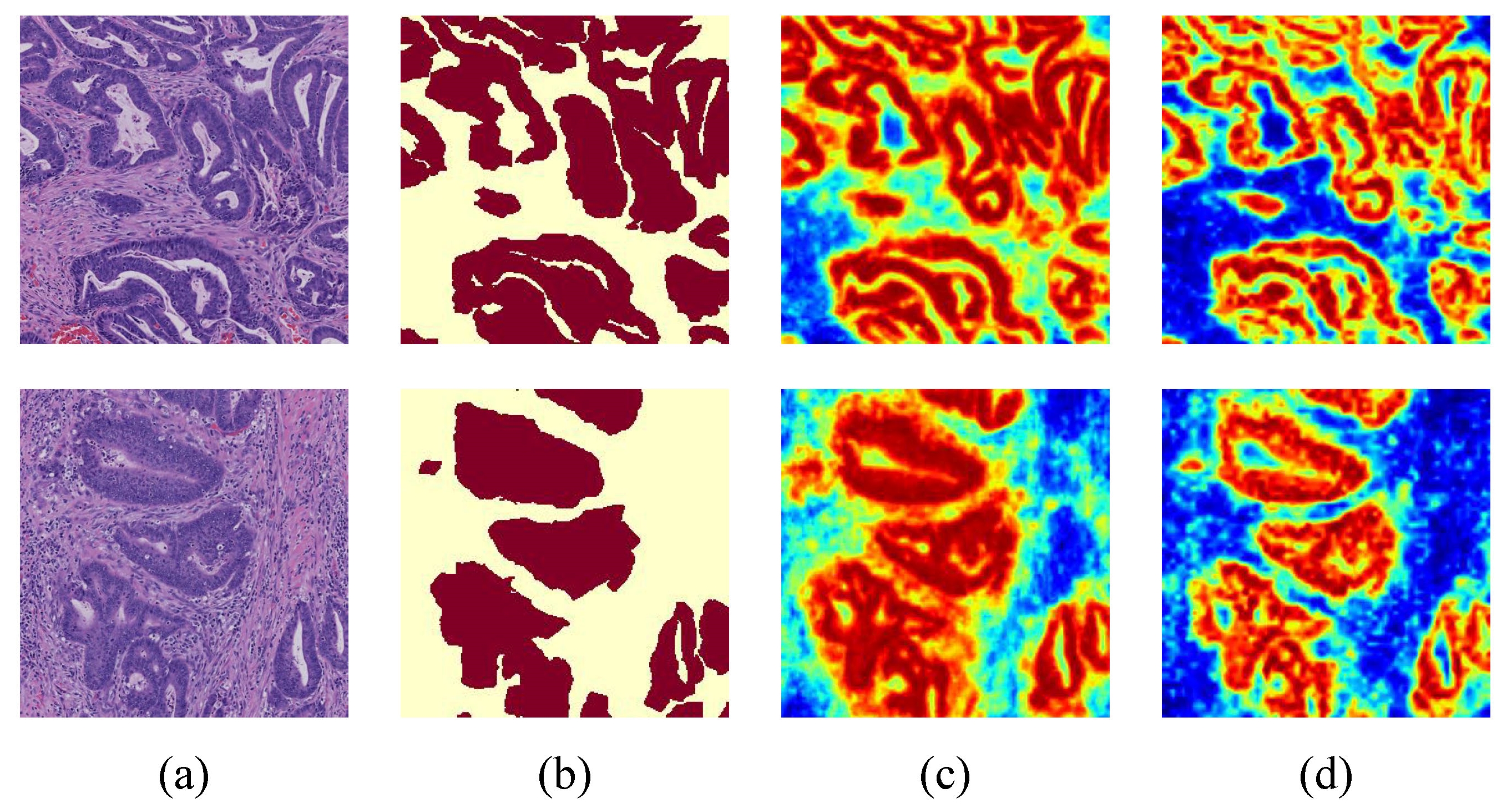}
\caption{Samples of feature maps of Transformer-based method and CNNs-based method. (a) Raw Image. (b) Ground Truth. (c) DWS-MIL. (d) Swin-MIL. }
\label{fig3}
\end{figure*}

\autoref{tab2} shows the comparison results adopting CNN model (VGG-16 \cite{simonyan2014very} \& ResNet18 \cite{targ2016resnet}) as the backbone. The VGG backbone can generate segmentation maps with higher boundary quality, while the ResNet backbone is able to extract more semantic information. However, because of the locality of convolution operations, the CNN backbones are unable to extract global or long-range dependencies, which can not solve the lack of relations between instances in MIL. By comparison, the Transformer backbone can learn global or long-range dependencies through multi-head self-attention, which leads to good performance both on F1-score (F1) and Hausdorff Distance (HD). Some samples of feature maps of Transformer-based method and CNNs-based method are shown in \autoref{fig3}.

\subsubsection{Effect of the number of stages in backbone:}
\autoref{tab3} summarizes the performances of the different number of stages in the Transformer backbone. It can be observed that the first two stages of Transformer backbone are unable to extract enough semantic information to perform segmentation accurately. On the other hand, the fourth stage incorporates semantic information more than necessary, which affects the performance of the boundary accuracy and fails to achieve effective improvements. Thus, taking the first three stages as our backbone is the optimal option.

\begin{minipage}{\textwidth}
\hspace{-10pt}
        \begin{minipage}[t]{0.45\textwidth}
        \centering

            \makeatletter\def\@captype{table}\makeatother\caption{\label{tab2}Effect of different backbone in MIL framework}
            \begin{tabular}{cccc}
\toprule
Backbone & $\rm F1^{Pos}$ & $\rm HD^{Pos}$ & $\rm F1^{Neg}$ \\ \hline
Swin-T & \textbf{0.850} & \textbf{10.463} & \textbf{0.999} \\
\;\;VGG-16\;\; & \;\;0.833\;\; & \;\;15.179\;\; & \;\;0.986\;\; \\
ResNet-18 & 0.838 & 17.852 & 0.948 \\ \bottomrule
\end{tabular}
        \end{minipage}
        \hspace{10pt}
        \begin{minipage}[t]{0.45\textwidth}
        \centering
        \makeatletter\def\@captype{table}\makeatother\caption{\label{tab3}Effect of the number of stages in backbone}
           \begin{tabular}{cccc}\toprule
Methods & $\rm F1^{Pos}$ & $\rm HD^{Pos}$ & $\rm F1^{Neg}$ \\ \hline
\;\;2 stages\;\; & \;\;0.814\;\; & \;\;12.980\;\; & \;\;0.998\;\; \\
3 stages & \textbf{0.850} & \textbf{10.463} & 0.999 \\
4 stages & 0.759 & 37.738 & \textbf{1.000} \\ \bottomrule
\end{tabular}
        \end{minipage}
    \end{minipage}

\subsubsection{Effect of side-outputs and fusion:}
To illustrate the effectiveness of deep supervision, \autoref{tab4} summarizes the performance of the different side-outputs. Due to the insufficient ability of feature extraction, the side-output of Stage 1 is limited in F1-score but gets the best boundary accuracy. The side-output of Stage 2 has the highest performance on F1-score and good boundary accuracy close to Stage 1. The side-output of Stage 3 contains most semantic information but lacks almost all boundary features. The fusion of three side-output feature maps can achieve optimal performance.

\begin{table}[h]
\caption{Effect of side-outputs and fusion}
\label{tab4}
\centering
\begin{tabular}{cccc}
\toprule
Methods & $\rm F1^{Pos}$ & $\rm HD^{Pos}$ & $\rm F1^{Neg}$ \\ \hline
\;\;\;Side-1\;\;\; & \;\;\;0.605\;\;\; & \;\;\;\textbf{10.383}\;\;\; & \;\;\;0.964\;\;\; \\
Side-2 & 0.810 & 10.413 & 0.995 \\
Side-3 & 0.743 & 100.965 & \textbf{1.000} \\
Fusion & \textbf{0.850} & 10.463 & 0.999 \\ \bottomrule
\end{tabular}
\end{table}
\section{Conclusion}
In this work, we propose a novel weakly supervised method for histopathology image segmentation, which combines Transformer with MIL to capture global or long-range dependencies. Multi-head self-attention in Transformer solves the independent instances of MIL by learning relations between instances, which significantly facilitates segmentation maps and makes the weakly supervised method more interpretable. It effectively utilizes hierarchical information from multi-scale feature maps through the introduction of deep supervision. The experiments demonstrate that our method achieves state-of-the-art performance on the colon cancer dataset. With a broad scope, the proposed method has the potential to be applied to a wide range of medical images in the future.
%
%
%
\bibliographystyle{splncs04}
\bibliography{ref}
\end{document}